\documentclass[11pt,a4paper]{article}
\usepackage[hyperref]{emnlp-ijcnlp-2019}
\usepackage{times}
\usepackage{latexsym}

\usepackage{url}
\usepackage[T1]{fontenc}
\usepackage[textsize=small]{todonotes}
\usepackage{amsmath}
\usepackage{amsfonts}
\usepackage{booktabs}
\usepackage{algorithm,algpseudocode,algorithmicx}
\usepackage{multirow}
\usepackage{rotating}
\usepackage{placeins}

\aclfinalcopy 

\title{Attending to Future Tokens for Bidirectional Sequence Generation}

\author{Carolin Lawrence \\
  Affiliation / Address line 1 \\
  Affiliation / Address line 2 \\
  Affiliation / Address line 3 \\
  \texttt{email@domain} \\\And
  Second Author \\
  Affiliation / Address line 1 \\
  Affiliation / Address line 2 \\
  Affiliation / Address line 3 \\
  \texttt{email@domain} \\}

\author{Carolin Lawrence \and Bhushan Kotnis \and Mathias Niepert \\
	NEC Laboratories Europe\\
	\texttt{\{carolin.lawrence, bhushan.kotnis, mathias.niepert\}@neclab.eu}
}

\date{}
\begin{document}
\maketitle
\begin{abstract}
	Neural sequence generation is typically performed token-by-token and left-to-right. Whenever a token is generated only previously produced tokens are taken into consideration. In contrast, for problems such as sequence classification, \emph{bidirectional} attention, which takes both past \emph{and} future tokens into consideration, has been shown to perform much better. We propose to make the sequence generation process bidirectional by employing special placeholder tokens. Treated as a node in a fully connected graph, a placeholder token can take past and future tokens into consideration when generating the actual output token. We verify the effectiveness of our approach experimentally on two conversational  tasks  where the proposed bidirectional model outperforms competitive baselines by a large margin.
\end{abstract}

\begin{figure*}
	\begin{center}
		\begin{tabular}{lccccccccccccc}
			\raisebox{.1\height}{\scriptsize 1} &$\bar{p}$&$\bar{p}$&$\bar{p}$&$\bar{p}$&$\bar{p}$&$\bar{p}$&$\bar{p}$&$\bar{p}$&$\bar{p}$&$\bar{p}$&$\bar{p}$&$\bar{p}$&$\bar{p}$\\
			\raisebox{.1\height}{\scriptsize 2} &$\bar{p}$& you& been& $\bar{p}$& or& enrolled& in& an& $\bar{p}$& degree& or& program& ?\\
			\raisebox{.1\height}{\scriptsize 3} &Have& you& been& accepted& or& enrolled& in& an& accredited& degree& or& program& ?\\
		\end{tabular}
		\caption{Example generation, going from a sequence of the placeholder token $\bar{p}$ (1), to an intermediate representation (2) and to the final output (3).}
		\label{tab:example}
	\end{center}
\end{figure*}

\section{Introduction}
When generating an output sequence, neural network models typically produce one token at a time. At each generation step, only the already produced sequence is taken into account. However, future and not-yet-produced tokens can also be highly relevant when choosing the current token. 
The importance of attending to both past and future tokens is apparent in self-attention architectures such as the Transformer~\citep{VaswaniEATL:17}. The self-attention module of a Transformer network treats a sequence \textit{bidrectionally} as a fully connected graph of tokens -- when a token is produced all other tokens are taken into consideration. However, this requires the entire sequence to be known a priori and when a Transformer is used for sequence generation, the self-attention process only includes previously produced tokens (\citet{VaswaniEATL:17,RadfordETAL:19}; \textit{inter alia}).
But the \textit{bidirectional} self-attention is a crucial property of the highly successful  language model BERT~\cite{DevlinETAL:18}. During the pre-training procedure of BERT, a fraction of input tokens is randomly masked out and the training objective is to predict these masked tokens correctly. BERT can then be fine-tuned for various classification tasks. Unfortunately, BERT cannot be directly used for sequence generation because the bidirectional nature of the approach requires the entire sequence to be known beforehand. 

Inspired by BERT's masking-based objective, we propose to start out with a sequence of placeholder tokens which are iteratively replaced by tokens from the output vocabulary to eventually generate the full output sequence. 
For an example see Figure \ref{tab:example}. With this novel model component, the self-attention of a Transformer can take both past and future tokens into consideration, leading to \textit{Bi}directional \textit{S}equence generati\textit{on} (\textsc{BiSon}). Furthermore, it allows us to directly incorporate the pre-trained language model BERT and, to the best of our knowledge, for the first time directly fine-tune it for sequence generation.

\textsc{BiSon} makes two major contributions which we investigate in turn. First, we explore different stochastic placeholder replacement strategies to determine, at training time, where to position the placeholder tokens. This is crucial as we need the \textsc{BiSon} models to be exposed to a large number of heterogeneous placeholder configurations. Second, we explore several strategies for iteratively generating, at inference time, a complete output sequence from an initial sequence of placeholders.

We evaluate our bidirectional sequence generation approach on two conversational tasks. \textsc{BiSon} outperforms both competitive baselines and state of the art neural network approaches on both datasets by a significant margin.

\section{Sequence Generation with Transformers}
For sequence-to-sequence tasks, an input sequence $x = x_1, x_2, \dots, x_{|x|}$ is to be mapped to an output sequence $y = y_1, y_2, \dots, y_{|y|}$ by some model $\pi_\theta$ with learnable parameters $\theta$. For neural models, this is typically done by first encoding the input sequence $x$ and then calling a decoder $t$ times to produce a sequence $y$ token-by-token, from left-to-right.

A popular choice for both encoder and decoder is the transformer \citep{VaswaniEATL:17}. It takes a sequence of embedded tokens $s = s_1, s_2, \dots, s_{|s|}$ and treats it as a fully connected graph over which a self-attention module is applied: for each token $s_t$ in the sequence it assigns a probabilistic attention score $a_t$ to every other token in the sentence. 
For the full mathematical details we refer the reader to \cite{VaswaniEATL:17}.

Typically a transformer \textit{encoder} is employed to encode $x$, whereas a transformer \textit{decoder} is used to produce $y$. In contrast to the encoder, the decoder at time step $t$ only has access to previously produced tokens $s_{<t} = s_1, s_2, \dots, s_{t-1}$. Consequently, the attention module cannot take possible future tokens into account when making its decision at time $t$. Additionally, in this encoder-decoder framework, there is a disconnect between input $x$ and output $y$ because the self-attention modules are applied to $x$ and $y$ in isolation before they are combined.

The latter weakness has been overcome in recent work \cite{RadfordETAL:18,WolfETAL:19,RadfordETAL:19} by feeding the concatenation $s = x \oplus y$ to a transformer decoder. At training time, given the current token $s_t$, the transformer is trained to predict the next word $s_{t+1}$ via maximum likelihood estimation. At test time, the transformer is conditioned on $x$ and then produces the output $y$ token-by-token. But because the model is a transformer decoder, it is unable to take possible future tokens into account.

\section{Bidirectional Sequence Generation}

During sequence generation, we want to take both past and future tokens into account. More formally, at time $t$, we want to attend to both $s_1, \dots, s_{t-1}$ as well as $s_{t+1}, \dots, s_{|s|}$. To do this, we give the sequence $s = x \oplus y$, the concatenation of the sequences $x$ and $y$, to a Transformer encoder, rather than a decoder. Of course, at inference time $y$ is unknown. Thus, we propose to replace each token $y_j$ with a \emph{placeholder token} $\bar{p}$. Since the model needs to be exposed to heterogeneous placeholder token configurations during training time, we introduce a \emph{placeholder strategy} that replaces some tokens $y_j$ with placeholder tokens $\bar{p}$ at training time. Hence, during training, a sequence $y$ is replaced by a sequence $p= p_1,p_2,…,p_{|y|}$, where a token $p_j$ is either the original token $y_j$ or the placeholder token $\bar{p}$. We introduce two placeholder strategies in the following section. At inference time, $p$ contains only placeholder tokens up to some pre-determined maximum sequence length.

With the placeholder strategy in place, a Transformer encoder is given the sequence $s = x \oplus p$ as input. The self-attention module then computes hidden representations $r_t$ of each token $s_t$ by attending to every other token in the sequence $s$. Because the output sequence is already present in the form of placeholder tokens both past tokens as well as future, not-yet-produced, tokens can be taken into consideration for every token $s_t$. 
Following the self-attention step, placeholder tokens are converted into a token from the output vocabulary with a language model (LM) classification layer, where for each placeholder $p_t$, its hidden representation $r_t$ is mapped to a distribution $d_t$ over the output vocabulary.

At training time, each output sequence token is fed the gold label and updates to the model $\pi_\theta$ are performed using stochastic gradient descent with a cross-entropy loss, i.e.

\[
\mathcal{L}_{\pi_\theta} =
- \frac{1}{M} \sum_{m=1}^{M} \sum_{j=1}^{|y|} \log \pi_\theta(p_j = y_j | s),
\]
where $M$ is the size of a minibatch.

At inference time, the placeholder tokens can be replaced iteratively based on the probability distribution $d_t$ over the output vocabulary for each placeholder $p_t$. Different sequence generation strategies are outlined in Section \ref{sec:generation}.

\subsection{Placeholder Replacement Strategy}\label{sec:placeholder}

At inference time, the output sequence starts out with a sequence of placeholder tokens. To introduce this notion at training time, we require a strategy that replaces some output sequence tokens $y_j$ with the placeholder token $\bar{p}$. The simplest approach would be to replace all output sequence tokens $y_j$ with the placeholder token $\bar{p}$. However, with this approach the model is never confronted with a sequence containing a mix of output sequence tokens and placeholder tokens.

Due to the exponential number of possible replacement configurations per given token sequence, we introduce probabilistic generative models that we can use to draw diverse sequence replacements. To this end, we model the decision whether to use placeholder or input tokens with probabilistic models, two of which we propose in the following. 

\paragraph{Bernoulli Random Variables (RV).}

We model each position of the input sequence with a binary random variable with a fixed mean. For every input sequence and every position $i$ in the sequence, we draw from a Bernoulli variable with mean $\mu$ to decide whether to use $y_i$ or $\bar{p}$. The expected number of placeholder tokens in a sequence of length $|y|$ is $|y| \mu$ and the variance is $\sigma^2 = |y| \mu(1-\mu)$. The variance of this strategy is determined by the mean and, therefore, the probabilistic model has one tunable parameter $\mu$. 

\paragraph{Gaussian Random Variables.}

The \emph{number of placeholder tokens} of the input sequence $p$ can be seen as drawn from an unknown optimal Binomial distribution. We can approximate this Binomial  with a normal distribution $\mathcal{N}(\mu, \sigma^2)$, where $\mu$ is the mean and $\sigma$ the standard deviation and they are considered hyperparameters. More formally, for every input sequence, we draw a value $P \sim \mathcal{N}(\mu, \sigma^2)$. Multiplied with the sequence length $|y|$, the nearest integer value $\lfloor (|y| \cdot P) \rceil$ is used as the number of placeholder tokens for the given sequence. The positions of the placeholder tokens in the sequence are then determined at random. 
The resulting probabilistic model's two parameters (mean $\mu$ and standard deviation $\sigma$) are treated as hyperparameters and are not updated during training. Being able to tune the variance parameter independently from the mean parameter might provide an advantage over the parameterization with Bernoulli RVs. 

\subsection{Sequence Generation Strategies}\label{sec:generation}

Starting with a sequence of placeholder tokens at inference time, it is possible to generate output token sequences in arbitrary order. We experiment with the following strategies. The distribution in all of these strategies are the distributions $d_t$ ($t=1,...,n$) for the placeholders over the output vocabulary. We use the term \emph{uncover} to mean that an output token is generated for a placeholder token. 

\textbf{One-step greedy.} In a single time step, all placeholder tokens are uncovered simultaneously by picking the most probable  token from the output vocabulary for each placeholder.

\textbf{Highest probability.} Placeholders are replaced iteratively and the placeholder to be uncovered is the placeholder that assigns the highest probability to a token from the output vocabulary, indicating the model is the most sure about this token.

\textbf{Lowest entropy.} Placeholders are replaced iteratively and the placeholder to be uncovered is the placeholder that exhibits the lowest entropy over its output vocabulary distribution and the most likely token at this position is chosen. Intuitively, the lowest entropy indicates the position where the uncertainty of the model to decide between tokens of the output vocabulary is the lowest.

\textbf{Left-to-right.} Placeholders are replaced iteratively, moving from left-to-right and thus mimicking the typical writing style for English. Note that this approach still differs from the Transformer decoders because future tokens are considered via the placeholder representations.

\textbf{No look ahead.} To test whether future placeholders hold useful information, we consider an adversarial sequence generation strategy: Again we iteratively uncover placeholders from left-to-right, but we suppress all attention flows from future placeholders. This imitates the behaviour of a transformer decoder but follows the idea of predicting a token on a placeholder, rather than predicting the next word as is typically done in transformer decoders. If this performs worse than left-to-right, there is indeed valuable information in future placeholder tokens.

\section{Experiments}

We conduct a series of experiments to explore \textsc{BiSon}'s behavior. First, we want to compare two token replacement strategies for training  as well as the four generation strategies for inference. Second, we want to compare \textsc{BiSon} to state of the art methods and investigate the impact of its ability to attend to future tokens. 

\subsection{Datasets}

We run experiments on the two following conversational datasets. 

\textbf{Goal-oriented \textsc{ShARC} \cite{SaeidiETAL:18}.}
\textsc{ShARC} is a dialogue, text-based question-answering dataset. Unlike many popular QA datasets, answers cannot simply be extracted from the text. Given a regulatory text, such as a text from the UK government's website, and a user scenario with corresponding question, it is necessary to interpret the text in the context of the specific user's needs. Before generating its final answer, a system may generate clarification questions. Finally, the system decides if the answer to the user's original question is ``\textit{Yes}'', ``\textit{No}'' or ``\textit{Irrelevant}'' where the latter means the question cannot be answered with the given text.

We perform the evaluation with the official \textsc{ShARC} script. For a set of generated clarification questions, it computes BLEU $n$-gram scores for $n={1, 2, 3, 4}$ using a set of clarification question in the set of gold responses. In each step of the conversation, the model under evaluation generates an output token sequence. This output is automatically assigned to the category ``\textit{More}'' if it is a clarification question, and to ``\textit{Yes}'', ``\textit{No}'', and ``\textit{Irrelevant}'' otherwise. Since this is a classification task we can compute micro and macro accuracy for it. The final model is chosen using the highest BLEU-4 score on the development set.

The \textsc{ShARC} dataset has a hidden test set and, therefore, it is not feasible to evaluate our various model variants. Hence, we take 30 unique rule texts and their corresponding training examples from the training set. This leads to a new development set of 2,465 instances and leaves the official development set to be used as a test set here. Finally we submitted our best model to be evaluated on the hidden test set.

\textbf{Free-form \textsc{Daily Dialog} \cite{LiETAL:17}.}
\textsc{Daily Dialog} is a dataset of written conversations occurring in daily life. Following the authors of the corpus, we report BLEU $n$-gram scores for $n={1, 2, 3, 4}$ for the generated output sequences with respect to the given gold responses. We tokenize these responses equally to ensure a fair comparison.

\subsection{\textsc{BiSon} Settings}
We implement \textsc{BiSon} based on the BERT Pytorch code\footnote{\url{https://github.com/huggingface/pytorch-pretrained-BERT}} and initialize with the pre-trained BERT model \textsc{bert-base-uncased} \cite{DevlinETAL:18}
. Consequently we employ the same model architecture and tokenisation as \citep{DevlinETAL:18} resulting in a model with about 110M parameters.  
To remain compatible with the BERT model, we prepend each sequence with a \textsc{[CLS]} token and place a \textsc{[SEP]} token after the input context. Similarly, producing a second \textsc{[SEP]} token indicates the end of sequence generation. For input context of \textsc{ShARC}, we follow \citet{SaeidiETAL:18} and use the concatenation of question, rule text, scenario and history. The input context for \textsc{Daily Dialog} is the concatenation of all previous utterances.

On the \textsc{ShARC} and \textsc{Daily Dialog} training sets we train for 20 and 40 epochs, respectively, which equates in each case to about $200k$ seen examples. 
As optimizer we used \textsc{Adam} \cite{adam} with $\beta_1 = 0.9$, $\beta_2 = 0.999$, a L2 weight decay of $0.01$ and a learning rate warm-up over the first 10\% of training steps. As learning rates we consider both the pre-training learning rate of BERT $1e$-$4$ and the fine-tuning learning rate $3e$-$5$. On preliminary experiments $3e$-$5$ proved to be best for \textsc{ShARC}, whereas it is $1e$-$4$ for \textsc{Daily Dialog}. We set the batch size to 15. Finally, the maximum sequence generation length, is set to 50 for \textsc{ShARC} and to 100 for \textsc{Daily Dialog}, which was chosen based on values observed in the training data. As the maximum sequence length of the BERT model is 512, longer input sequences are truncated accordingly. For the main results, we employ the sequence generation strategy left-to-right, which we found to work best. Later on we also report results for the other strategies.

For the Bernoulli RV approach, we test $\mu \in [0.2,0.8]$ with increments of $0.1$. For the Gaussian RV approach, we test all possible combinations for the following hyperparmeters $\mu = \{0.4, 0.5, 0.6\}$ and $\sigma = \{0.3, 0.6, 0.9\}$. The best combination on the \textsc{ShARC} dev set is $\mu = 0.5, \sigma = 0.6$. It outperforms the best Bernoulli approach ($\mu = 0.7$) by 3.4 point in BLEU-4 score. Some Bernoulli experiments in fact only produced a very small number of clarification question, e.g. $\mu = 0.5$ only generated 9 clarification questions on the development set, whereas in the ground truth responses 846 clarification questions occur.
This suggests that a high variance is important, as the Bernoulli setups all have a variance of $0.25$ or lower and our best Gaussian approach has a variance of $0.6$.
We directly employ the Gaussian distribution with $\mu = 0.5, \sigma = 0.6$ on the \textsc{Daily Dialog} task. 

\subsection{Baselines}
To measure the success of our proposed approach, we consider the following three baselines.

\textbf{Encoder-Decoder Transformer (E\&D).}
First, we compare our bidirectional encoder to a standard encoder-decoder Transformer where the decoder only has access to tokens produced so far to compute its self-attention. 
We use the implementation of OpenNMT \cite{opennmt} and employ the parameters suggested by them, but adjust the learning rate to $0.1$, which we found to work better for both datasets. Additionally, we increased the word and hidden dimension size to 768 and the number of attention heads to 12 to match the capacity of our model. Training ran for 50 epochs. Needing both an encoder and a decoder, this leads to a total of about 270M parameters.

\textbf{Encoder-Decoder Transformer with BERT (E\&D+B).}
The power of our bidirectional decoder stems from two advantages. First, we can initialize our model with the pre-trained \textsc{bert-base-uncased} model. Second, the decoding process is bidirectional. It would be possible to transfer the first advantage to an encoder-decoder framework by using BERT embeddings. This is however only possible for the input sequence, because the bidirectionality of BERT requires the entire sequence to be available beforehand. 
Thus, we modify implementation of OpenNMT to use the BERT model as the encoder. The weights are frozen when training the decoder, which produced better results than allowing the gradients to also flow through the BERT model. Again, with both an encoder and decoder, this leads to a total of about 270M parameters.

\textbf{GPT2.}
\citet{RadfordETAL:19} present a transformer decoder, GPT2, trained as a language model on large amounts of monolingual text. \citet{RadfordETAL:19} showed that it is possible to perform various tasks in a zero-shot setting by priming the language model with an input and letting it generate further words greedily. This setup can be transferred to a supervised setting, where the model is fine-tuned to a dataset by using maximum likelihood estimation to increase the probability of the gold output sequence \citep{WolfETAL:19}. As the starting point for the supervised learning, we initialize the model with the pre-trained model \textsc{GPT-2-117M} released by \citet{RadfordETAL:19}\footnote{\url{https://github.com/openai/gpt-2}} and then fine-tune. With 117M parameters, this model is comparable to our model.
Unlike baseline 2, this setup can directly employ a pre-trained model as our approach can, but it is not bidirectional.
\begin{table}
	\begin{center}
		\begin{tabular}{ll*{4}{p{0.9cm}}}
			\toprule
			&Model& Micro Acc. & Macro Acc. & B-1 & B-4\\
			\midrule
			\multirow{4}{*}{\begin{sideways}\textsc{ShARC}\end{sideways}}&E\&D&31.9&38.9&17.1&$\hphantom{0}$1.9\\
			\cmidrule{2-6}
			&E\&D+B&54.7&60.4&24.3&$\hphantom{0}$4.3\\
			&GPT2&60.4&65.1&53.7&33.9\\
			&\textsc{BiSon}&\textbf{64.9}&\textbf{68.8}&\textbf{61.8}&\textbf{46.2}\\
			\bottomrule
		\end{tabular}
		\caption{Results on the \textsc{ShARC} test set, averaged over 3 independent runs for GPT2 and \textsc{BiSon}, reporting micro accuracy and macro accuracy in terms of the classification task and BLEU-1 and BLEU-4 on instances for which a clarification question was generated. E\&D uses no language model pre-training.}
		\label{tab:sharc}
	\end{center}
\end{table}
\begin{table}
	\begin{center}
		\begin{tabular}{ll*{4}{p{0.9cm}}}
			\toprule
			&Model& Micro Acc. & Macro Acc. & B-1 & B-4\\
			\midrule
			\multirow{2}{*}{\begin{sideways}\scriptsize\textsc{ShARC}\end{sideways}}&\textsc{E3}&\textbf{67.6}&\textbf{73.3}&54.1&38.7\\
			&\textsc{BiSon}&66.9&71.6&\textbf{58.8}&\textbf{44.3}\\
			\bottomrule
		\end{tabular}
		\caption{Results on the official hidden \textsc{ShARC} test set of our model compared to the best model on the leaderboard, E3 \cite{ZhongZettlemoyer:19}.}
		\label{tab:sharc_leader}
	\end{center}
\end{table}

\subsection{Results}
We report the results of our approach, the various baselines, as well as the previous state-of-the-art (SOTA) scores where applicable in Table \ref{tab:sharc} and \ref{tab:sharc_leader} for \textsc{ShARC} and in Table \ref{tab:daily} for \textsc{Daily Dialog}.

On the \textsc{ShARC} dataset, we observe very poor BLEU-4 performance for the encoder-decoder Transformer (E\&D), which is consistent with results from \citet{SaeidiETAL:18}, who could not get a LSTM-based network to work without an additional classification head. Adding BERT (E\&D+B) slightly improves performance. By directly leveraging a pre-trained model, GPT2 outperforms the previous models by a large margin, reaching 33.9\% on BLEU-4 and a micro accuracy of 60.4\%. \textsc{BiSon} is able to take future tokens into consideration and outperforms GPT2 by 12.3 percentage points in BLEU-4 and by 4.5 points in micro accuracy. 

We submitted the best \textsc{BiSon} model out of the random three of Table \ref{tab:sharc} to be evaluated on the hidden test set and report results in comparison to the best model on the leaderboard,\footnote{\url{https://sharc-data.github.io/leaderboard.html}, 19 August 2019} \textsc{E3} \citep{ZhongZettlemoyer:19} in Table \ref{tab:sharc_leader}. \textsc{BiSon} outperforms E3 by 5.6 BLEU-4 points, while it is only slightly worse than E3 in terms of accuracy.

On the \textsc{Daily Dialog} dataset the information retrieval-based method (IR in Table \ref{tab:daily}) introduced by \citet{LiETAL:17} is very strong and outperforms the best end-to-end model (E2E) \citep{LuoETAL:18} by over 16 percentage points in BLEU-4. The best end-to-end model is based on LSTMs and \citet{LuoETAL:18} report performance increases when adding an attention module to their setup. The encoder-decoder transformer (E\&D) outperforms this setup by over 2 percentage points in BLEU-4 and we conjecture that this is due to the transformer making more effective use of the attention principle. Adding BERT (E\&D+B) does not help much for this dataset. But again we observe a large increase of performance when directly employing pre-trained models. GPT2 performs on par with the IR SOTA, achieving a BLEU-4 score of 19.4\%. Again, \textsc{BiSon} can outperform GPT2, here with a difference of 6.2 points in BLEU-4 and even larger increases in the other scores.

\begin{table}
	\begin{center}
		\begin{tabular}{ll*{4}{p{0.9cm}}}
			\toprule
			&Model&B-1 &B-2&B-3& B-4\\
			\midrule
			\multirow{6}{*}{\begin{sideways}\textsc{\textsc{Daily Dialog}}\end{sideways}}&IR&-&25.8&20.4&19.4\\
			&E2E&14.2&5.7&3.8&$\hphantom{0}$2.8\\
			&E\&D&22.3&6.8&5.7&$\hphantom{0}$5.2\\
			\cmidrule{2-6}
			&E\&D+B&26.1&7.3&6.0&$\hphantom{0}$5.5\\
			&GPT2&42.3&23.6&20.7&19.4\\
			&\textsc{BiSon}&\textbf{54.9}&\textbf{32.6}&\textbf{28.0}&\textbf{25.6}\\
			\bottomrule
		\end{tabular}
		\caption{BLEU $n$-gram scores for $n={1, 2, 3, 4}$ on the DailyDialog test set, averaged over 3 independent runs for GPT2 and \textsc{BiSon}. Models before the line do not make use of a pre-trained language model. IR (SOTA) \cite{LiETAL:17} and E2E (SOTA) \cite{LuoETAL:18} are, to the best of our knowledge, the best previously published scores for information retrieval and end-to-end approaches.}
		\label{tab:daily}
	\end{center}
\end{table}

\textbf{Effect of bidirectionality.}
To investigate that our model benefits from bidirectionality, we consider a setup where \textsc{BiSon} isn't allowed to attend to future tokens during prediction (see Table \ref{tab:no_look_ahead}). It causes a drop in BLEU-4 performance of about 25 points on the \textsc{ShARC} dataset and a drop of 10 points on the \textsc{Daily Dialog} dataset. This showcases that \textsc{BiSon} during training has learnt to rely on the ability to attend to future tokens.

\begin{table}
	\begin{center}
		\begin{tabular}{ll*{4}{p{0.88cm}}}
			\toprule
			&Model& Micro Acc. & Macro Acc. & B-1 & B-4\\
			\midrule
			\multirow{2}{*}{\begin{sideways}\scriptsize\textsc{ShARC}\end{sideways}}&\textsc{BiSon}&\textbf{64.9}&\textbf{68.8}&\textbf{61.8}&\textbf{46.2}\\
			&past only&64.3&67.4&35.0&21.3\\
			\midrule
			\midrule
			&Model&B-1 &B-2&B-3& B-4\\
			\midrule			
			\multirow{2}{*}{\begin{sideways}\textsc{DD}\end{sideways}}&\textsc{BiSon}&\textbf{54.9}&\textbf{32.6}&\textbf{28.0}&\textbf{25.6}\\
			&past only &48.0&24.6&18.5&14.8\\
			\bottomrule
		\end{tabular}
		\caption{Comparison of \textsc{BiSon} to a setup where \textsc{BiSon} isn't allowed to attend to future tokens, i.e. past only, for \textsc{ShARC} and \textsc{Daily Dialog} (DD).}
		\label{tab:no_look_ahead}
	\end{center}
\end{table}
\textbf{Effect of pre-trained model.}
We are curious how big the effect of the pre-trained model is. Thus, instead of starting with the \textsc{bert-base-uncased} weights, we initialize \textsc{BiSon} with random weights drawn from a normal distribution with mean 0.0 and standard deviation of 0.02. 
Results are presented in Table \ref{tab:pre-train} for \textsc{ShARC} and \textsc{Daily Dialog}. Even without a pre-trained language model, our approach can outperform the standard encoder-decoder transformer framework (E\&D) on both datasets, although we had to increase the number of epochs for the \textsc{ShARC} dataset to 40. On the \textsc{Daily Dialog} task, we are even able to outperform GPT2. This demonstrates the effectiveness of our approach in itself, free of any pre-trained language model.

\begin{table}
	\begin{center}
		\begin{tabular}{ll*{4}{p{0.9cm}}}
			\toprule
			&Model& Micro Acc. & Macro Acc. & B-1 & B-4\\
			\midrule
			\multirow{2}{*}{\begin{sideways}\scriptsize\textsc{ShARC}\end{sideways}}&E\&D&31.9&38.9&17.1&$\hphantom{0}$1.9\\
			&\textsc{BiSon}&\textbf{52.9}&\textbf{57.4}&\textbf{21.9}&$\hphantom{0}$\textbf{2.3}\\
			\midrule
			\midrule
			&Model&B-1 &B-2&B-3& B-4\\
			\midrule			
			\multirow{2}{*}{\begin{sideways}\textsc{DD}\end{sideways}}&E\&D&22.3&6.8&5.7&$\hphantom{0}$5.2\\
			&\textsc{BiSon}&\textbf{46.3}&\textbf{27.0}&\textbf{23.6}&\textbf{22.4}\\
			\bottomrule
		\end{tabular}
		\caption{Best end-to-end models that do not use a pre-trained language model in comparison with \textsc{BiSon} that uses randomly initialized weights for \textsc{ShARC} and \textsc{Daily Dialog} (DD), averaged over 3 runs.}
		\label{tab:pre-train}
	\end{center}
\end{table}

\begin{table}
	\begin{center}
		\begin{tabular}{lcc}
			\toprule
			Strategy& \textsc{ShARC} & \textsc{Daily Dialog}\\
			\midrule
			one step greedy&22.9&$\hphantom{0}$9.3\\
			lowest entropy&40.3&16.8\\
			highest probability&\textbf{50.9}&16.4\\
			left-to-right&46.2&\textbf{23.8}\\
			\bottomrule
		\end{tabular}
		\caption{BLEU-4 using various sequence generation strategies for \textsc{BiSon} on  \textsc{ShARC} and \textsc{Daily Dialog}.}
		\label{tab:predict}
	\end{center}
\end{table}

\textbf{Effect of sequence generation strategies.}
We present the different sequence generation strategies in Table \ref{tab:predict}. The best overall sequence generation strategy is to predict from left to right which achieves good results on both datasets. On the \textsc{ShARC} dataset the highest probability approach performs better than left-to-right. However, on \textsc{Daily Dialog} this approach is not as successful. 
This suggests that it might be worth selecting the best sequence generation strategy for each dataset individually. However, we hypothesize that left-to-right works consistently well due to the left-to-right nature of the English language. A brief experiment with a right-to-left strategy gave poor results.

\section{Analysis}\label{sec:analysis}

\begin{table}
	\begin{center}
		\begin{tabular}{lccc}
			\toprule
			Dataset& $\alpha^1$ & $\alpha^2$ & $\alpha^3$\\
			\midrule
			\textsc{ShARC} &92.6$\pm3.1$&5.2$\pm2.4$&2.2$\pm1.8$\\
			DD &97.0$\pm2.5$&2.3$\pm2.3$&0.7$\pm0.4$\\
			\midrule
			\midrule
			&& $\bar{\alpha}^2$ & $\bar{\alpha}^3$\\
			\midrule
			\textsc{ShARC} &-&71.7$\pm13.7$&28.3$\pm13.7$\\
			DD &-&70.2$\pm14.5$&29.8$\pm14.5$\\
			\bottomrule
		\end{tabular}
		\caption{Average attention weights and standard deviation when predicting from left-to-right on both \textsc{ShARC} and \textsc{Daily Dialog} (DD) for different parts of the sequence, where $\alpha^1$ is for the input sequence $x$, $\alpha^2/\bar{\alpha}^2$ is for the already produced sequence $y$ and $\alpha^3/\bar{\alpha}^3$ is for the sequence of remaining placeholder tokens $p$. $\alpha^k$ are the normalized attention weights across all three parts, whereas $\bar{\alpha}^k$ normalizes over the second and third part.}
		\label{tab:attention}
	\end{center}
\end{table}

We believe that the placeholders capture sequential information present in the language model learned during pre-training. After running a transformer encoder where each position can attend to every other position, a placeholder token will have a probability distribution over the output vocabulary and this distribution is informed by all other tokens in input and output. Thus, a placeholder could be seen as a mixture of tokens with varying probabilities. As placeholders are subsequently uncovered, the other placeholders can update their distribution by taking the newly revealed token into consideration.

\begin{figure}
	\centerline{\includegraphics[width=0.45\textwidth,keepaspectratio]{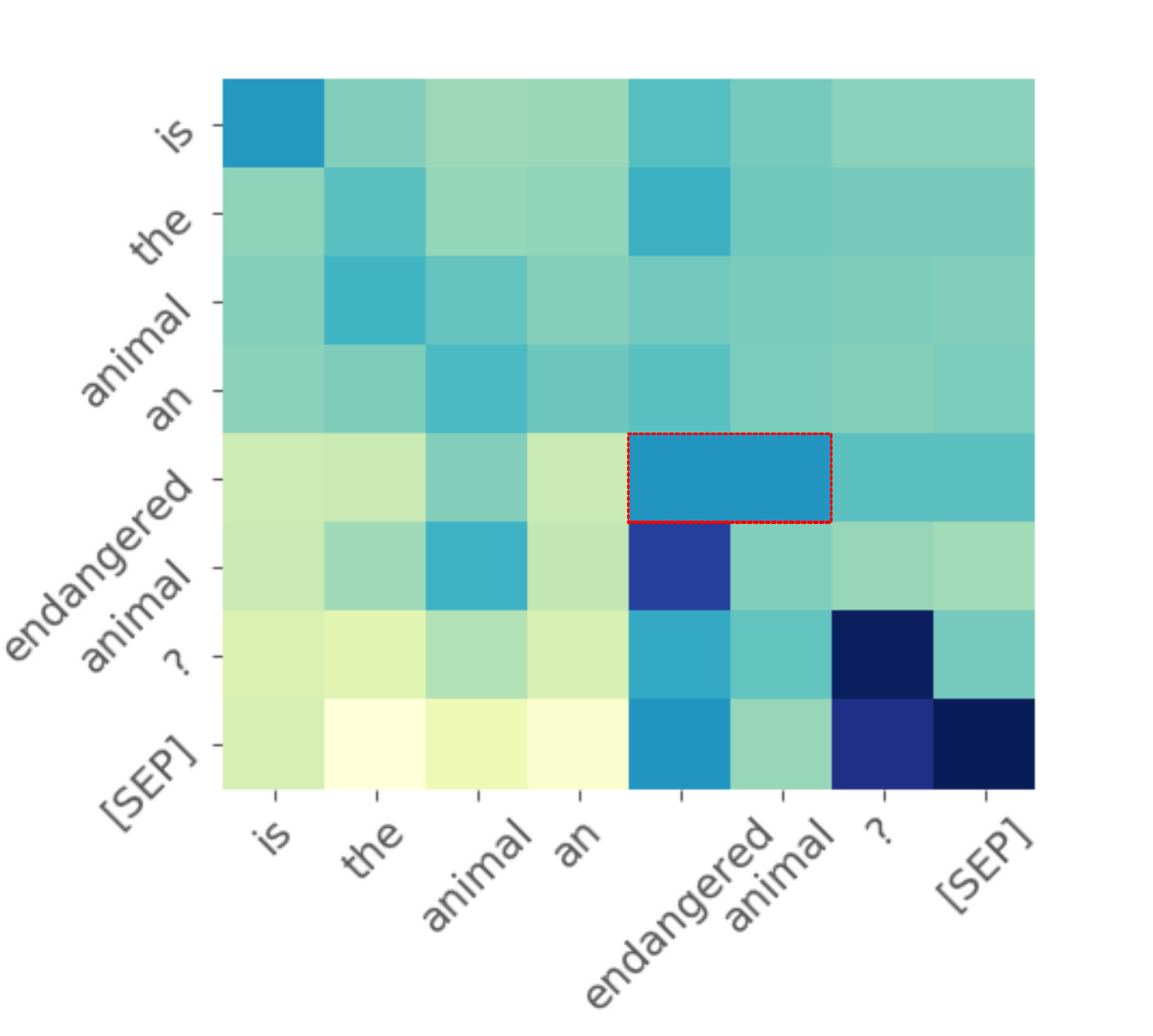}}
	\caption{Heat map (darker hues indicate higher attention) that shows an example of where an attention head looks into the future while generating from left to right. Each row shows the attention over the output sequence for this row's placeholder token at that point in time. Word in previous rows have been produced already, whereas words of later rows still hold placeholder tokens. Thus the upper triangle of the matrix shows the attention that is paid to future tokens. The red square shows that while generating the token ``endangered'', the attention head already takes the next placeholder into account, which is revealed to be ``animal'' in the next step. Best viewed in color.}
	\label{fig:heatmap1}
\end{figure}

For example, in Figure \ref{fig:heatmap1}, for the sentence``is the animal an endangered animal ?'', while generating ``endangered'', the self-attention head pays attention to the next placeholder token, which in the next step is revealed to be ``animal''. While producing ``endangered'', the distribution for the next position already placed a high probability on ``animal'', thus the current token can take this into consideration and produces ``endangered''. Further heat maps demonstrating this can be found in the appendix.

The quantify this intuition, we measure the average attention score on various parts of the sequence. For this, we use the left-to-right prediction strategy. Thus, at time $t$, we can decompose our sequence into three parts: $s = x \oplus y \oplus p$, where $x$ is the input, $y$ the already produced sequence and $p$ the remaining sequence of placeholder tokens. For each attention head, we can decompose the attention probabilities into the three parts, 
\vspace{-2.mm}
\begin{enumerate}
	\setlength\itemsep{0.2mm}
	\item attention on the input text,
	\item attention on the current word and already generated words (left of the current word),
	\item attention on words that are yet to be generated (right of the current word).
\end{enumerate}
This is mathematically expressed as
\begin{align*}
a_t &= a_{0:|x|} \oplus a_{|x|+1:|x|+t} \oplus a_{|x|+t+1,|s|}, \\ \notag
\end{align*}

where $|s|$ is the maximum possible sequence length. For each part we can calculate an average leading to three values, $a_t^1, a_t^2$ and $a_t^3$.

Averaged over all $T$ generation time steps and all $D$ data points, we can derive a score for each part $k,\ k = 1,2,3$ and each attention head $h$:

\begin{align*}
\alpha_h^k = \frac{1}{D}\frac{1}{T} \sum_{d=1}^{D} \sum_{t=1}^{T} a_{d,t}^k
\end{align*}
Note that we use the attention heads in the final BERT self-attention layer. Averaging over all $H$ attention heads, $\alpha^k = \frac{1}{H} \sum_{h=1}^{H}  \alpha_h^k$, leads to the results reported in Table  \ref{tab:attention} for both datasets.
Unsurprisingly, we find that  with scores of over 90\% for both datasets the majority of the attention is focused on the first part, i.e. the conditioning input $x$ (see $\alpha^1$ in Table \ref{tab:attention}). The remaining attention is split between the already produced sequence ($\alpha^2$) and the future tokens ($\alpha^3$).

To directly compare the relationship within the sequence generation, we re-normalize over $\alpha^2$ and $\alpha^3$, leading to new values $\bar{\alpha}^2$ and $\bar{\alpha}^3$ (see Table \ref{tab:attention}). Here we can see that the past, already produced tokens are about twice as important as the future, not-yet-produced tokens. But with scores of just under 30\% on both datasets, we see that a substantial amount of attention is also focused on the future, not-yet-produced tokens. 

Interestingly, with a standard deviation of about 14\%, the values of $\bar{\alpha}^2$ and $\bar{\alpha}^3$ vary strongly across the different attention heads. For example on the \textsc{ShARC} dataset, we find one attention head where only about $9\%$ is focused on the future and another where it is about $64\%$ and thus this attention head pays more attention to the future than the past. A graphical overview can be found in the appendix for both datasets.

\section{Related Work}
Transformers \cite{VaswaniEATL:17} model sequences as fully connected graphs and apply a bidirectional self-attention module where every token can attend to every other token. Because of this a Transformer is not restricted to sequential orderings. However, \citet{VaswaniEATL:17}; \textit{inter alia} still restrict themselves to producing tokens from left-to-right and only allow a Transformer decoder to attend to previously produced tokens. Recently, several attempts have been made to lift the left-to-right restriction in Transformer or LSTM-based models \citep{GuETAL:19,SternETAL:19,WelleckETAL:19,LongETAL:19}, but in those approaches it is not possible to attend to future, not-yet-produced tokens.

Concurrently to our work, \cite{GhazvininejadETAL:19} proposed a similar placeholder strategy approach for generating in the context of machine translation. However, they employ an encoder-decoder framework, whereas we only require an encoder, which more closely links input and output via a single shared attention module. Furthermore, they only consider uniform sampling of placeholders whereas we found that the higher variance, which we can control with the Gaussian random variable approach, leads to better results. 

Bidirectionality is one of the crucial ingredients in the success of the recently proposed unsupervised language model BERT \cite{DevlinETAL:18}. 
For this, \citet{DevlinETAL:18} propose a Transformer \textit{encoder} to take full advantage of the bidirectional nature of the Transformer. Their resulting model, BERT, can directly be applied to various classification tasks but not to sequence generation tasks. Our approach shows how a Transformer encoder can be used for sequence generation and this allows us to directly incorporate BERT into our experiments. 

GPT \citep{RadfordETAL:18} and GPT2 \citep{RadfordETAL:19} are both pre-trained language models that use a Transformer \textit{decoder} instead, which can only attend to already produced tokens. 
For dialogue, the GPT model has been fine-tuned for the chit-chat dataset PersonaChat \cite{ZhangETAL:18} by \citet{WolfETAL:19}. 
While GPT and GPT2 can immediately be used as a sequence generators, these models do not offer bidirectionality and they cannot attend to not-yet-produced tokens. Our bidirectional encoder for sequence generation can combine the best of both worlds.

\section{Conclusion}
We introduced bidirectional sequence generation by employing placeholders in the output sequence. These placeholder tokens are subsequently replaced by tokens of the output vocabulary. Crucially, this allows a transformer encoder to attend to both past and future, not-yet-produced token. Simply masking all placeholder tokens is not feasible. Instead we investigated two placeholder strategies, based on Bernoulli and Gaussian random variables. 
At prediction time, our approach is not restricted to produce the output sequence from left to right. However, this strategy proved to produce most consistent results in our experiments.

Our approach outperforms previous end-to-end approaches that do not make use of any pre-trained language models. In conjunction with the pre-trained language model BERT, our bidirectional sequence generation approach allows us to achieve new state-of-art results on both conversational tasks.
In the future, we would like to apply our approach to other sequence generation tasks. Additionally, we wonder if a further performance increase could be achieved if the pre-training of BERT would employ our placeholder strategy.

\bibliography{lit}
\bibliographystyle{acl_natbib}

\appendix
\section*{Appendix}
\section{Forward \& Backward Attention}
\begin{figure}[h]
	\centerline{\includegraphics[width=0.5\textwidth,keepaspectratio]{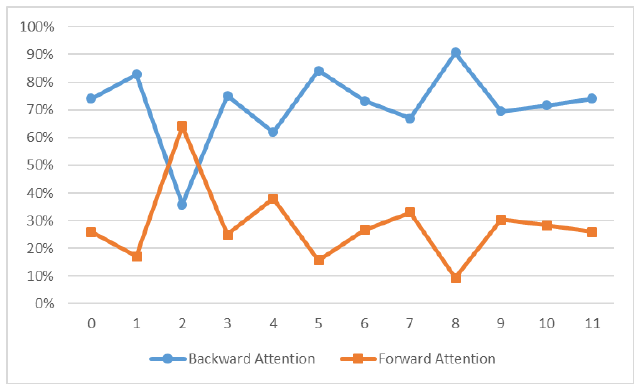}}
	\caption{$\bar{\alpha}^2$ (Backward Attention) and $\bar{\alpha}^3$ (Forward Attention) across the 12 attention heads for the \textsc{ShARC} dataset.}
	\label{fig:attention_sharc}
\end{figure}

\begin{figure}[h]
	\centerline{\includegraphics[width=0.5\textwidth,keepaspectratio]{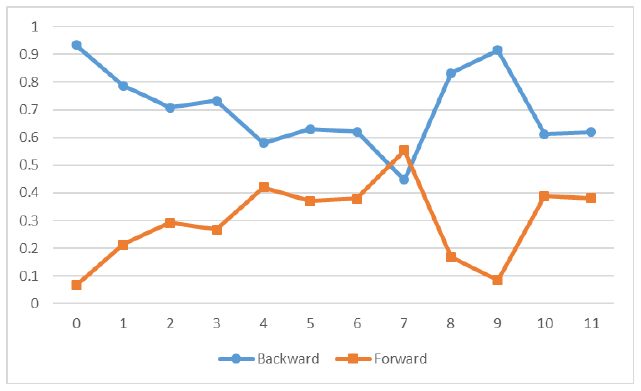}}
	\caption{$\bar{\alpha}^2$ (Backward Attention) and $\bar{\alpha}^3$ (Forward Attention) across the 12 attention heads for the \textsc{Daily Dialog} dataset.}
	\label{fig:attention_dd}
\end{figure}

Figures \ref{fig:attention_sharc} and \ref{fig:attention_dd} present the normalized forward attention $\bar{\alpha}^2$ and backward attention $\bar{\alpha}^3$ for the different attention heads over the sequence generation part for the \textsc{ShARC} and \textsc{Daily Dialog} dataset, respectively.

\FloatBarrier

\FloatBarrier

\section{Heat Maps}
\begin{figure}[h]
	\centerline{\includegraphics[width=0.5\textwidth,keepaspectratio]{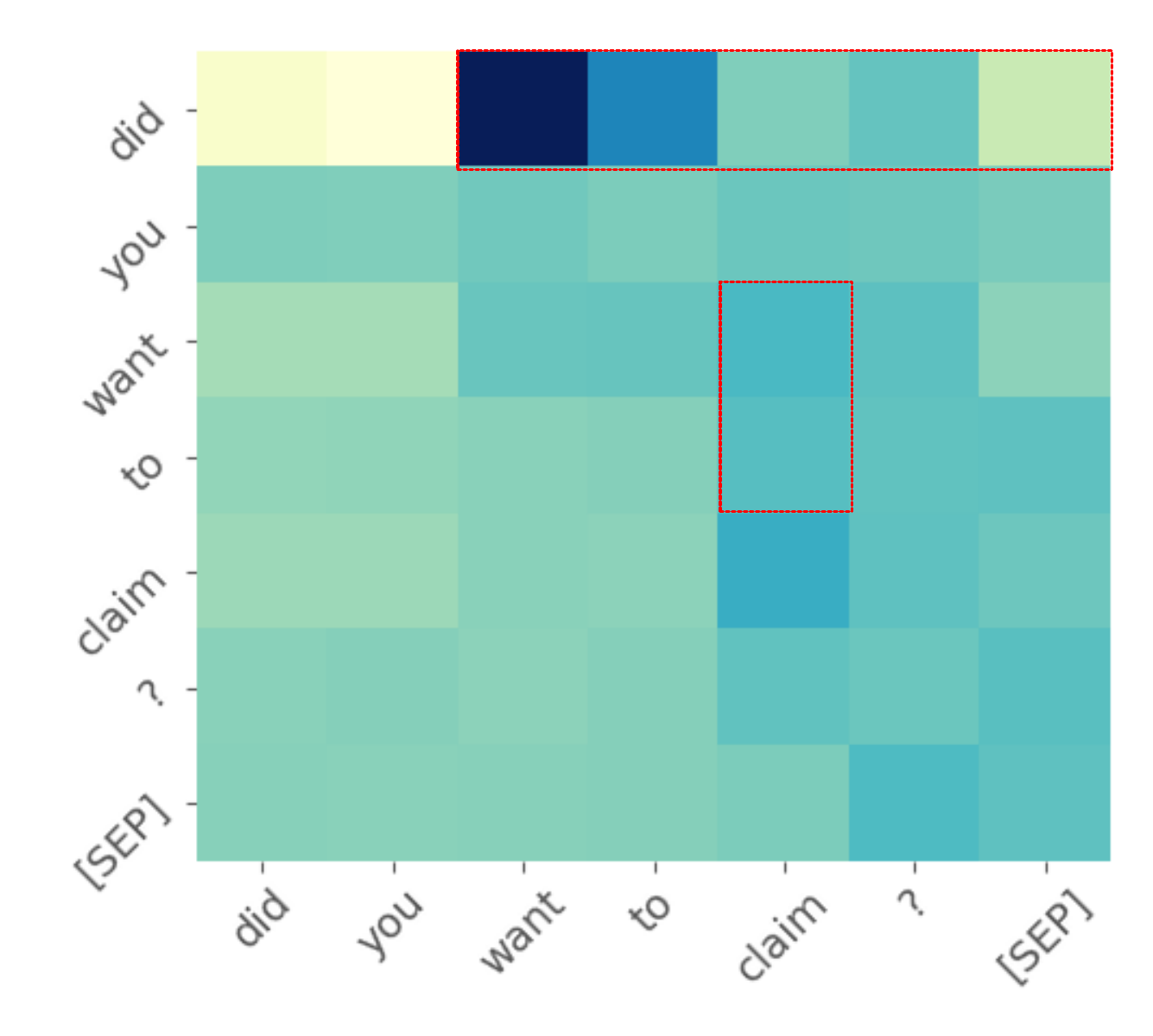}}
	\caption{}
	\label{fig:heatmap2}
\end{figure}

\begin{figure}[h]
	\centerline{\includegraphics[width=0.5\textwidth,keepaspectratio]{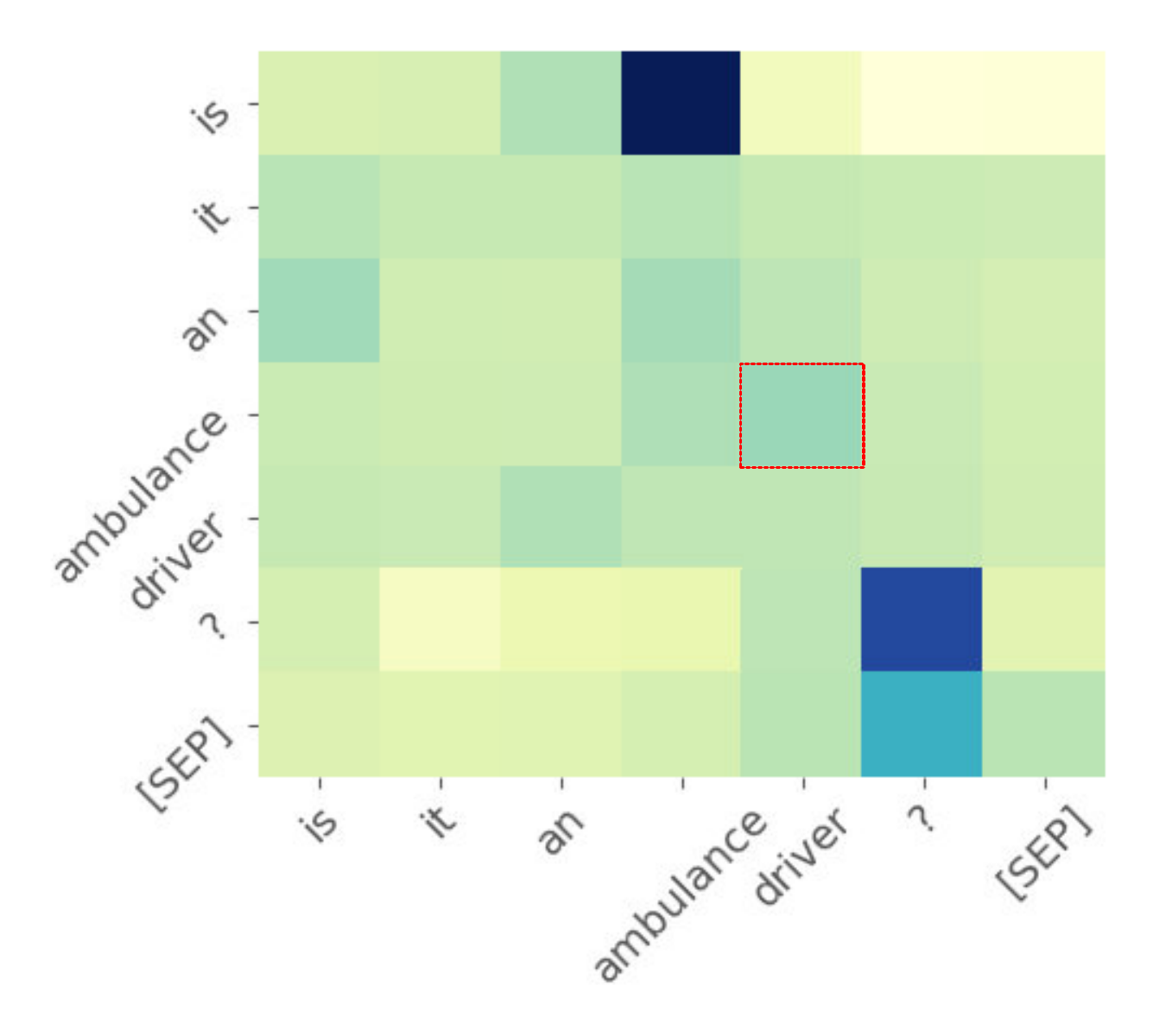}}
	\caption{}
	\label{fig:heatmap3}
\end{figure}

The heat maps (darker hues indicate higher attention) of Figures \ref{fig:heatmap2} and \ref{fig:heatmap3} show examples of where an attention head strongly looks into the future while generating from left to right. Each row shows the attention over the output sequence for this row's placeholder token at that point in time. Word in previous rows have been produced already, whereas words of later rows still hold placeholder tokens. Thus the upper triangle of the matrix shows the attention that is paid to future tokens. The red square marks the point of interest. Best viewed in color.

In Figure \ref{fig:heatmap2}, we can see that when deciding on the first word, ``\textit{did}'', high attention is paid to important future tokens. In particular, there is a strong focus on the 3rd placeholder, which is later revealed to be ``\textit{want}''. Attention is also paid to the position that is later revealed to become a question mark. This indicates that model plans ahead and realizing a question will be formed, the word ``\textit{did}'' is chosen. Note that the similar sentence, ``you want to claim.'' would not require the word ``\textit{did}'' as it would not be a question.

Also in Figure \ref{fig:heatmap2}, both the words ``\textit{want}'' and ``\textit{to}'' pay strong attention to the final word ``\textit{claim}'' in the phrase ``\textit{want to claim}''.

In Figure \ref{fig:heatmap3}, when producing the word ``\textit{ambulance}'' the attention is focused on the next placeholder token, which is in the next step revealed to be the word ``\textit{driver}'' in the noun phrase ``\textit{ambulance driver}''.

\end{document}